\documentclass[journal]{IEEEtran}
\ifCLASSINFOpdf
\else
\fi
\usepackage{times}
\usepackage{epsfig}
\usepackage{graphicx}
\usepackage{amsmath}
\usepackage{amssymb}
\usepackage{caption}
\usepackage{multicol}
\usepackage{breqn}
\usepackage{booktabs}
\usepackage{tabularx}
\usepackage{multirow}
\usepackage{xcolor,colortbl}
\definecolor{Gray}{gray}{0.85}
\newcolumntype{a}{>{\columncolor{Gray}}c}

\newcommand{\se}[1]{\textcolor{blue}{#1}}
\newcommand{\fir}[1]{\textcolor{red}{#1}}

\begin{document}
%
\title{Visualizing the Invisible: Occluded Vehicle Segmentation and Recovery}
%
%
%

\author{Xiaosheng Yan,
        Yuanlong Yu,
        Feigege Wang,
        Wenxi Liu,
        Shengfeng He,
        and Jia Pan,
}

%
%

\markboth{Journal of \LaTeX\ Class Files,~Vol.~14, No.~8, August~2015}%
{Shell \MakeLowercase{\textit{et al.}}: Bare Demo of IEEEtran.cls for IEEE Journals}
%



\maketitle

\begin{abstract}
In this paper, we propose a novel iterative multi-task framework to complete the segmentation mask of an occluded vehicle and recover the appearance of its invisible parts. In particular, to improve the quality of the segmentation completion, we present two coupled discriminators and introduce an auxiliary 3D model pool for sampling authentic silhouettes as adversarial samples. In addition, we propose a two-path structure with a shared network to enhance the appearance recovery capability. By iteratively performing the segmentation completion and the appearance recovery, the results will be progressively refined.
To evaluate our method, we present a dataset, the Occluded Vehicle dataset, containing synthetic and real-world occluded vehicle images. We conduct comparison experiments on this dataset and demonstrate that our model outperforms the state-of-the-art in tasks of recovering segmentation mask and appearance for occluded vehicles. Moreover, we also demonstrate that our appearance recovery approach can benefit the occluded vehicle tracking in real-world videos. 
\end{abstract}


%
\IEEEpeerreviewmaketitle

\section{Introduction}

In recent years, the semantic segmentation and instance segmentation techniques have made significant progress due to the development of deep learning~\cite{long2015fully,yu2016multi,chen2018deeplab,he2017mask,dai2016instance,chen2018masklab,liu2018path}.
Despite the achieved impressive performance, it is still difficult to accurately reason about objects under occlusions in a two-dimensional image. On the contrary, according to the study on amodal perception~\cite{kanizsa1979organization}, one main strength of the human visual system is the ability to reason about the invisible, occluded parts of objects with high fidelity. 
To reduce the gap between the vision models and the human visual system, recent works start to investigate the problem of inferring the invisible part of objects, including amodal instance segmentation~\cite{li2016amodal,zhu2017semantic} and generating the invisible part of objects~\cite{ehsani2018segan}.

\begin{figure}
	\hspace{.2cm}
	\footnotesize
	\setlength{\tabcolsep}{3pt}
	\begin{tabular}{cc}
		\includegraphics[width=0.22\textwidth,trim={0 1.3cm 0 2.6cm},clip]{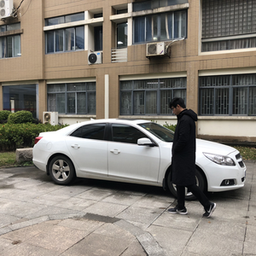}&
		\includegraphics[width=0.22\textwidth,trim={0 1.3cm 0 2.6cm},clip]{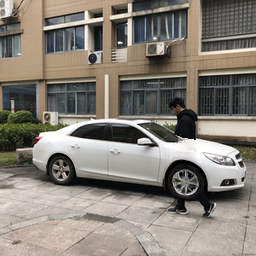}\\
		{\footnotesize (a) Input image with an occluded car}  & {\footnotesize (b) Recovered appearance}\\
		\includegraphics[width=0.22\textwidth,trim={0 2cm 0 2.3cm},clip]{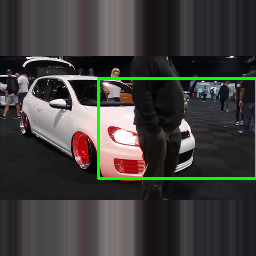}&
		\includegraphics[width=0.22\textwidth,trim={0 2cm 0 2.3cm},clip]{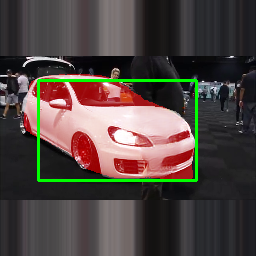}\\
		\includegraphics[width=0.22\textwidth,trim={0 1.6cm 0 2.6cm},clip]{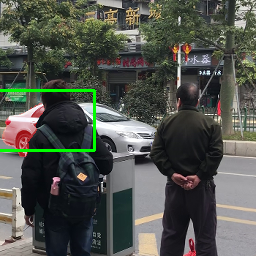}&
		\includegraphics[width=0.22\textwidth,trim={0 1.6cm 0 2.6cm},clip]{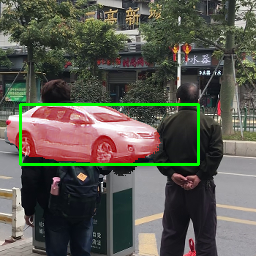}\\
		{\footnotesize (c) Tracking in original videos}  & {\footnotesize (d) Tracking in recovered videos}\\
	\end{tabular}
	\caption{\small (a-b) Given the input image with an occluded car, our approach can recover the appearance of its invisible part. Illustration of tracking occluded vehicles in the original video (c) and our processed video (d) that recovers the appearance of the target vehicles from occlusions. }  
	\vspace{-0.3cm}
	\label{fig:teaser} 
\end{figure}

In this paper, we focus on the task of appearance recovery for occluded vehicles. As we know, identifying vehicles is crucial for the applications of visual surveillance, intelligent traffic control, path prediction, and autonomous driving. However, in scenarios with vehicles and pedestrians, the occlusions are often observed and they increase the difficulty of learning visual representation of vehicles. As shown in Fig.~\ref{fig:teaser}, the tracker may fail to follow the target under occlusions, since the occlusions coexist with the foreground target, which prevents the tracker from learning the accurate features of the target. In this situation, recovering the appearance of the invisible parts will mitigate such problem and benefit the tracking. 


To accomplish the appearance recovery, we propose an iterative multi-task framework of segmentation completion and appearance recovery for occluded vehicles. 
Our framework consists of two modules: a segmentation completion network that aims at completing the incomplete segmentation mask of the occluded vehicle and an appearance recovery network that aims to recover its appearance. 
In particular, to accurately recover the segmentation of the occluded vehicles, we propose two coupled discriminators, i.e., one object discriminator and one instance discriminator, in the segmentation completion network (see Fig.~\ref{fig:framework1}). The instance discriminator encourages the network to generate a segmentation mask similar to the ground-truth of the instance, while the object discriminator forces the produced mask to be similar to a real vehicle.
To accomplish this, we introduce an auxiliary 3D model pool to generate adversarial samples. Although the rendered 3D models are visually different from the real cars, their silhouettes are authentic compared with the real ones and thus they are suitable as the adversarial samples to further improve the generation quality. Since the types and poses of 3D vehicle models can be of a wide variety, this pool of 3D models implicitly brings richer prior into the segmentation completion. 

In addition, to generate the visible parts from occlusions, we propose a two-path architecture as the appearance recovery network (see Fig.~\ref{fig:framework2}). On the training stage, one path learns to fill in the colors of the invisible parts, while the other path is assigned with a more challenging task for inpainting the entire foreground vehicle given the image context. 
Since the parameters of networks on both paths are shared during training, the capability of the appearance recovery network will be enhanced. At test time, only the first path is deployed for generation. 
Lastly, our proposed multi-task framework allows the recovered image to be processed multiple times for refinement. 
To evaluate our method, we present an Occluded Vehicle dataset (OVD) that contains synthetic and real images.
We test our approach on this dataset by comparing with the state-of-the-art methods in tasks of segmentation completion and appearance recovery. Moreover, we also apply our approach to recover the occluded vehicles in several real-world video sequences and demonstrate that our recovery approach can benefit the tracking as well.

The contributions of this paper are summarized as:
\begin{itemize}
	\item We propose a novel iterative multi-task framework that consists of two sequential modules to obtain the complete segmentation of an occluded vehicle and recover its appearance. 
	\item To infer the complete segmentation of a vehicle, we propose a segmentation completion network which has two coupled discriminators, which integrates an auxiliary 3D model pool to generate adversarial samples.
	\item To recover the appearance of the occluded vehicles, we present a two-path network architecture, which incorporates a foreground inpainting task with the appearance recovery of the invisible parts on the training stage to improve the recovery quality.
	\item We present a dataset, called the Occluded Vehicle Dataset (OVD), containing synthesized and real images of occluded vehicles for training and validation. Based on this dataset, we demonstrate that our work outperforms the state-of-the-art methods. Besides, we also collect several video sequences to demonstrate our approach can benefit occluded vehicle tracking.
\end{itemize}
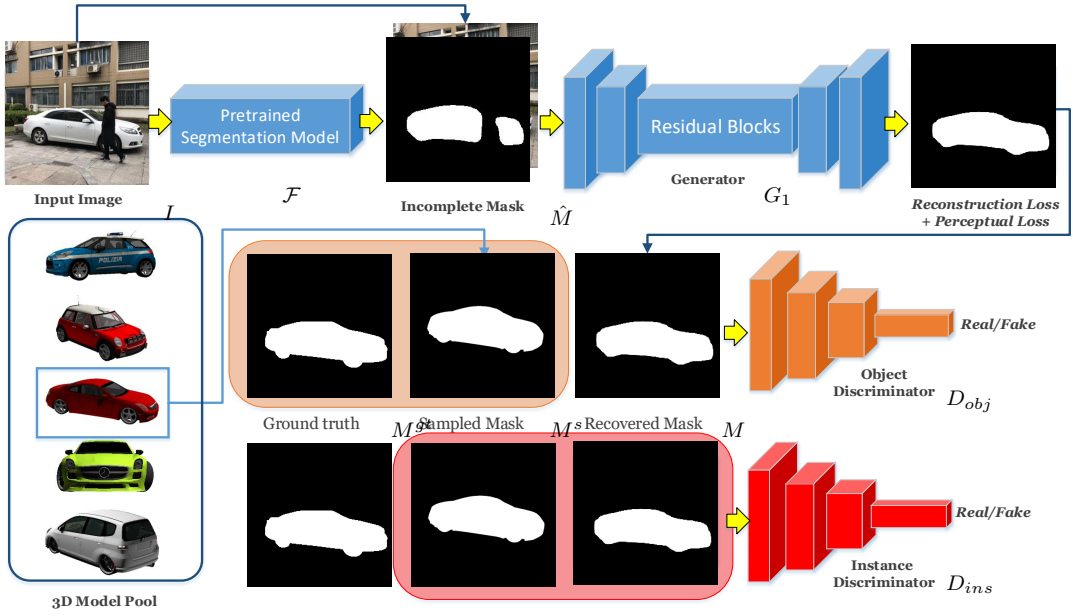
\begin{figure*}
	\centering
	\includegraphics[width=0.8\textwidth]{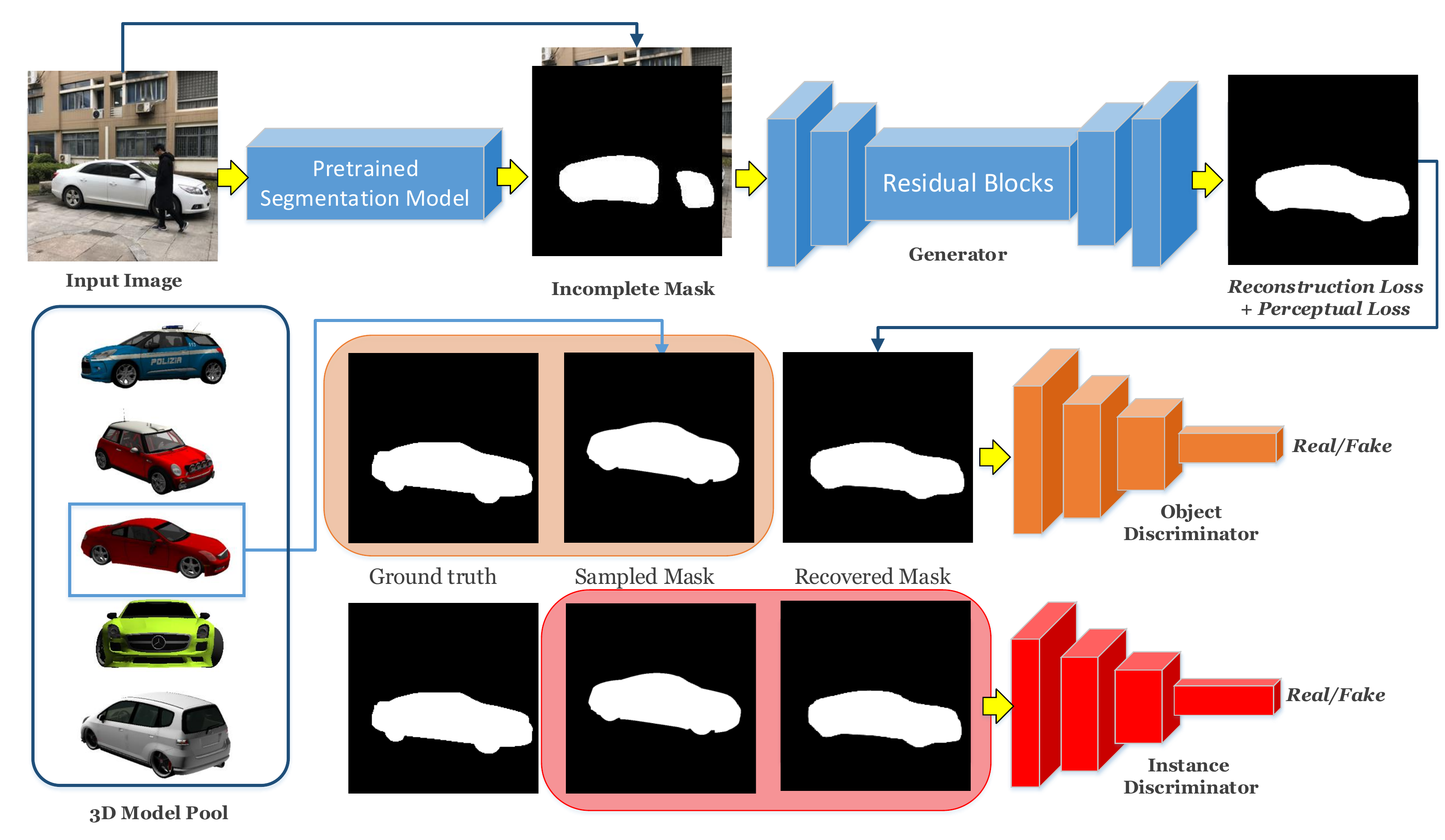}
	\put(-300,159){\footnotesize $\mathcal{F}$}
	\put(-118,159){\footnotesize $G_1$}
	\put(-345,152){\footnotesize $I$}
	\put(-199,150){\footnotesize $\hat{M}$}
	\put(-259, 70){\footnotesize $M^{gt}$}
	\put(-199, 70){\footnotesize $M^s$}
	\put(-134, 70){\footnotesize $M$}
	\put(-49,  82){\footnotesize $D_{obj}$}
	\put(-49,  12){\footnotesize $D_{ins}$}
	\vspace{-0.3cm}
	\caption{\small Illustration of the segmentation completion network. The input image $I$ is passed to the pretrained segmentation model $\mathcal{F}$ and then concatenated with the computed incomplete mask $\hat{M}$ to produce the recovered segmentation mask $M$. In our framework, we present two coupled discriminators, both of which are fed with the same samples for different classification tasks. For the object discriminator $D_{obj}$, it aims to categorize the ground-truth $M^{gt}$ and the sampled silhouette mask $M^{s}$ as real and the recovered mask $M$ as fake. For the instance discriminator $D_{ins}$, it aims to classify the ground-truth $M^{gt}$ as real, while to classify the sampled silhouette mask $M^{s}$ and the recovered mask $M$ as fake. Specifically, the silhouette mask $M^{s}$ is sampled from the 3D model pool as an extra adversarial sample. 
	}
	
	\label{fig:framework1}
	
\end{figure*}

\section{Related Works}

We survey the related literature on occlusion handling, generative adversarial network, and vehicle related works.

\textbf{Occlusion handling.} 
The occlusions are often observed in images or videos, which is often challenging in many vision problems.
Thus, there are prior works studying the occlusion reasoning \cite{yang2012layered,gao2011segmentation,tighe2014scene,chen2015multi,hsiao2012occlusion}. It has also been extensively studied in the detection and tracking community~\cite{koller1994robust,yang2005real,shu2012part,zhang2014partial,hua2014occlusion,mueller2017real}, but these works do not consider recovering the appearance of the occluded objects. 
On the other hand, the amodal segmentation problem has been specifically presented and studied by ~\cite{li2016amodal,zhu2017semantic,ehsani2018segan,follmann2018learning}, which aim at providing a complete mask for occluded objects. 
Among the prior works, most similar to ours is Ehsani et al.~\cite{ehsani2018segan}, which presents SeGAN to generate the invisible parts of objects from indoor scenes.
In their model, they deploy a residual network to produce a completed segmentation mask, which is too trivial to fully learn and recover the mask of the occluded objects with various shapes and poses. Besides, the resolution of their produced segmentation mask is much lower than that of the input image, which degrades its performance.
Unlike SeGAN, we present an improved GAN model with two coupled discriminators to generate high-quality masks with the assistance of the silhouette masks sampled from various 3D models as adversarial samples. 

\textbf{Generative adversarial network.}
Generative adversarial network (GAN) is composed of a generative model and a discriminative model competing against each other in a two-player min-max game. It has been extensively studied~\cite{goodfellow2014generative,chen2016infogan,arjovsky2017wasserstein} and widely applied in many applications, e.g. image-to-image translation~\cite{isola2017image,zhu2017unpaired}. Besides, GAN has been applied in image inpainting~\cite{pathak2016context,yu2018generative,yu2016multi}. SeGAN~\cite{ehsani2018segan} also adopts GAN to generate the appearance of the invisible parts. However, their model requires the previously recovered segmentation mask as the only input, and thus it heavily relies on the quality of the input segmentation mask and is lack of image contextual information. In our work, we present a two-path architecture integrated with an inpainting task that allows the network to fully learn from the image context.

\textbf{Vehicle related works.}
There are extensive studies on vehicles in the vision community, including detection~\cite{sivaraman2013looking,yan2017exploiting}, tracking~\cite{mei2011robust}, counting~\cite{zhang2017fcn}, and re-identification~\cite{shen2017learning,wang2017orientation,zhouy2018viewpoint}. In addition, with the rapid advancement in autonomous driving, more related research topics have been investigated~\cite{kim2017interpretable,behl2017bounding,ramanishka2018toward,maqueda2018event}. However, there are only a few recent works focusing on occluded vehicles, including vehicle detection under occlusion~\cite{zhang2008multilevel,ma2017detection,chang2018vision} and vehicle tracking with occlusions~\cite{pang2004novel,zhao2016appos}. Different from prior works, our work focuses on image-based occluded vehicle appearance recovery.

\section{Our Proposed Framework}

\subsection{Overview}

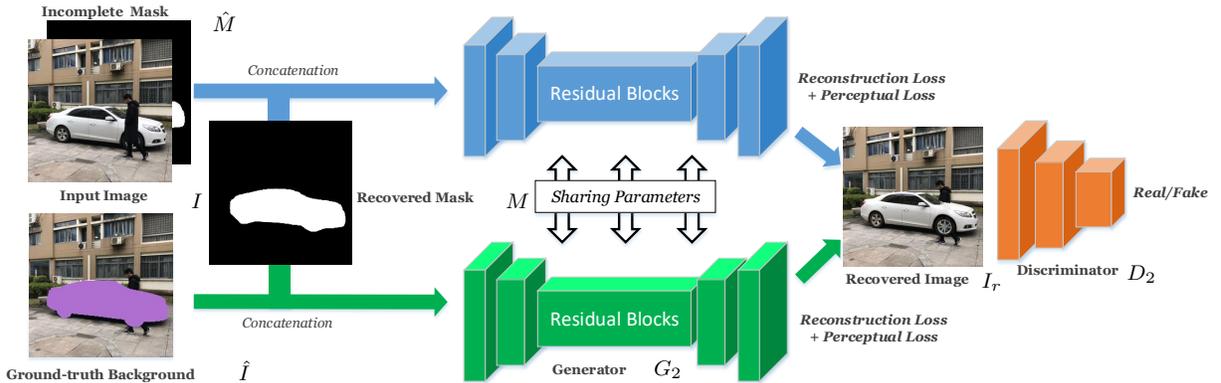
\begin{figure*}[t]
	\centering
	\includegraphics[width=0.9\textwidth]{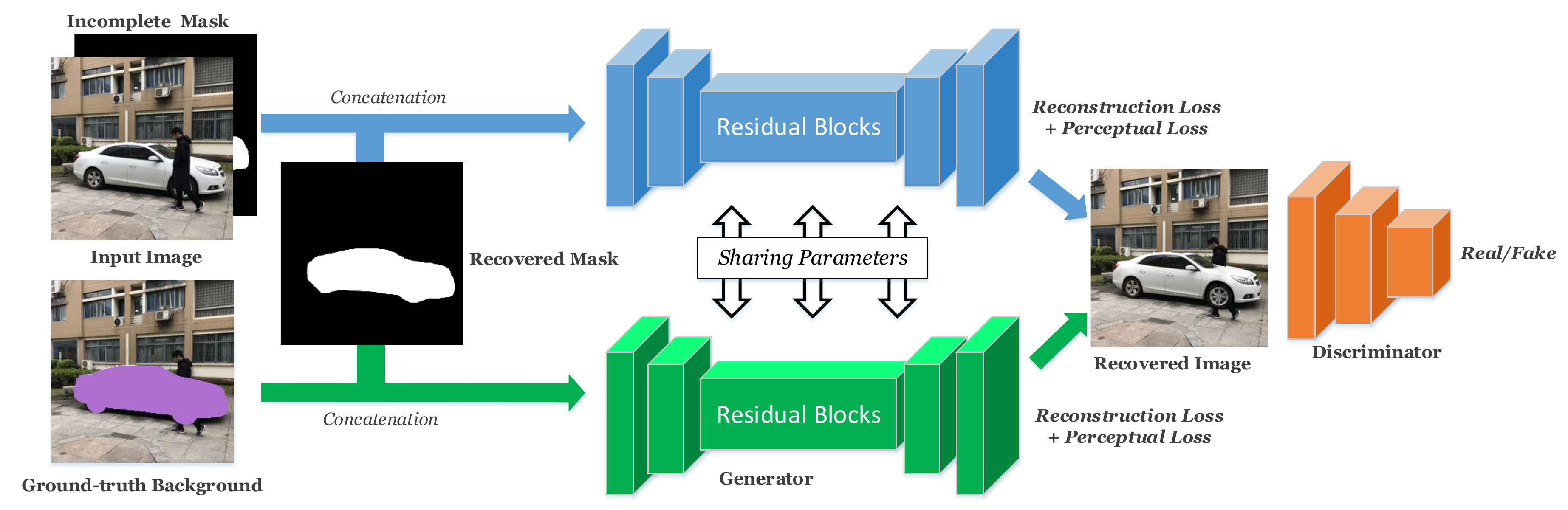}
	\put(-213,  7){\footnotesize $G_2$}
	\put(-388, 70){\footnotesize $I$}
	\put(-380,137){\footnotesize $\hat{M}$}
	\put(-370,  5){\footnotesize $\hat{I}$}
	\put(-269, 70){\footnotesize $M$}
	\put(-34,  43){\footnotesize $D_2$}
	\put(-89,  40){\footnotesize $I_r$}
	\vspace{-0.3cm}
	\caption{\small Illustration of the appearance recovery network. On the training stage, its generator has two paths to perform separate generation tasks but shares the same network. The first path aims at filling in colors for invisible parts.
		The second path aims at inpainting the image foreground given the image culling out the foreground object. At test time, only the first path is adopted for generating appearance. 
	}
	\label{fig:framework2}
	
\end{figure*}

Our framework is composed of two networks: the segmentation completion network (Fig.~\ref{fig:framework1}) and the appearance recovery network (Fig.~\ref{fig:framework2}). In particular, given the input image containing an occluded vehicle, the segmentation completion network generates the recovered segmentation mask. Then, the recovered segmentation mask is passed through the 
appearance recovery network to produce the invisible parts of vehicles. After painting the invisible parts back to the original image, the occluded vehicle will be on the foreground of the image. Lastly, the image will be fed through the segmentation completion network and the appearance recovery network multiple times to enhance the recovery quality.
In the following subsections, we will introduce the details of both networks.

\subsection{Segmentation Completion Network}

As illustrated in Fig.~\ref{fig:framework1}, the segmentation completion network is based on a GAN model.
In specific, the input image $I$ containing occluded vehicles is first fed into a pretrained segmentation model $\mathcal{F}$ to generate the incomplete mask.
The incomplete segmentation mask, $\hat{M}$, is then concatenated with the input image $I$ and then passed into an encoder-decoder, i.e. the generator $G_1$. Its loss is the combination of the reconstruction loss (i.e. $\mathcal{L}_1$ loss) and the perceptual loss. 

In a standard GAN, the generated sample and its corresponding ground-truth will be passed to a discriminator for classification,
but the standard GAN tends to produce coarse results.
To improve the quality of produced segmentation mask, we present two coupled discriminators including an object discriminator $D_{obj}$ and an instance discriminator $D_{ins}$. In particular, the instance discriminator, similar to the standard discriminator, is used to force the network to generate a mask identical with the ground truth. The object discriminator, on the other hand, aims to encourage the network to generate a mask similar to a real vehicle. 

To accomplish this, we introduce an auxiliary 3D model pool, which collects a variety of rendered 3D vehicle models and their corresponding accurate silhouettes from ShapeNet~\cite{shapenet2015}. 
Although the rendered 3D models are visually different from the real cars, their silhouettes are similar to the real ones. Besides, it is easy to extract accurate contour masks from the rendered images of these models.
Thus, they are suitable as the adversarial samples to improve the generation quality. 
By randomly selecting a silhouette as the adversarial sample, we collect three types of masks for discrimination, i.e. the ground-truth mask $M^{gt}$, the recovered mask $M$, and the sampled silhouette $M^{s}$.
Inspired by the discriminator design of \cite{zhang2018stackgan}, as illustrated in Fig.~\ref{fig:framework1}, the object discriminator aims to classify whether its input masks are real vehicle masks or not, i.e., to categorize the ground-truth and the sampled silhouette as real, and the recovered mask as fake, which is formulated as:
\begin{align}
\mathcal{L}_{adv}(G_1, D_{obj}) = \mathbb{E}_{\hat{M}}[\log (1-D_{obj}(G_{1}(I,\hat{M}))] + \nonumber\\  \frac{1}{2}\left(\mathbb{E}_{M^{gt}}[\log D_{obj}(M^{gt})] + \mathbb{E}_{M^{s}}[\log D_{obj}(M^{s})]\right),
\end{align}
The purpose of the instance discriminator is to classify whether the input mask is the segmentation of the vehicle, i.e., to categorize the ground-truth as real, but the sampled silhouette and the recovered mask as fake, which is defined:
\begin{align}
\mathcal{L}_{adv}(G_1,{D_{ins}}) = \mathbb{E}_{M^{gt}}[\log D_{ins}(M^{gt}, I, \hat{M})] + \nonumber\\
\frac{1}{2}(\mathbb{E}_{\hat{M}}[\log (1-D_{ins}(G_{1}(I,\hat{M}), I, \hat{M}))]+\nonumber\\
\mathbb{E}_{M^{s}}[\log (1-D_{ins}(M^{s}, I, \hat{M}))]).
\end{align}
Note that $D_{ins}$ uses the image $I$ and mask $\hat{M}$ as additional inputs to attentively focus on the visible part of the occluded vehicle for discrimination.
During training, a variety of different vehicles silhouettes are sampled, which encourages the network to learn the general feature of real vehicle masks (e.g. the location and shape of wheels), and thus to encourage the generation of the masks similar as real vehicles. 
Hence, the final objective of the segmentation completion network is to minimize:
\begin{align}
\mathcal{L}^{seg} = \mathcal{L}_{adv}(G_1, D_{obj})+\mathcal{L}_{adv}(G_1, D_{ins})+ \nonumber\\
\lambda \mathcal{L}_{L1}(G_1) + \beta \mathcal{L}_{perc}(G_1),
\end{align}
\noindent where $\mathcal{L}_1(\cdot)$ and $\mathcal{L}_{perc}(\cdot)$ denote the reconstruction loss and the perceptual loss, respectively.

\subsection{Appearance Recovery Network}

With the recovered segmentation mask $M$, our framework aims to generate the appearance of the invisible parts for the occluded vehicle in the next step. As illustrated in Fig.~\ref{fig:framework2}, the appearance recovery network is also based on a GAN model.

As the generator, we propose a two-path architecture which performs two separate tasks while sharing the same network $G_2$. For the first path, the recovered segmentation mask $M$ is integrated with the input image $I$ and the incomplete mask $\hat{M}$, to fed into the first path. Since the incomplete mask indicates the visible parts and the recovered mask estimates the silhouette of the whole unoccluded vehicle, the purpose of this path is to fill in the colors for invisible parts. 

In addition, the second path is supervised to perform a more challenge task, i.e. to inpaint the whole vehicle based on image context.
It receives the concatenation of the recovered mask $M$ and the ground-truth background $\hat{I}$ along with a zero map $\phi$ for padding in order to learn for inpainting. 
When training, the network $G_2$ is shared on both paths, so it will be endowed with the ability that not only recovers the invisible parts but also the entire vehicle based on contextual information, which significantly improves the capability of the generator (See Tab.~\ref{tab:ab1}).
The recovered images are sent to the discriminator $D_2$ to guarantee the image quality. Thus, objective of the appearance recovery network is to minimize:
\begin{align}
\mathcal{L}^{app} = &\mathcal{L}_{adv}(G_{2}(I,\hat{M},M), D_2)+\mathcal{L}_{adv}(G_{2}(\hat{I},M,\phi), D_2)+ \nonumber\\
&\lambda_1 \mathcal{L}_{L1}(G_{2}(I,\hat{M},M)) + \beta_1 \mathcal{L}_{perc}(G_{2}(I,\hat{M},M))+ \nonumber\\
&\lambda_2 \mathcal{L}_{L1}(G_{2}(\hat{I},M,\phi)) + \beta_2 \mathcal{L}_{perc}(G_{2}(\hat{I},M,\phi)).\nonumber
\end{align}
On the testing stage, since the ground-truth background of the test image is unknown, the second path is disabled and the generator on the first path is applied. 

\subsection{Iterative Refinement}

On the testing stage, the recovered image can be produced by passing the input image $I$ through both generators, i.e., $I_r = G_2(G_1(\mathcal{F}(I)))$, where $\mathcal{F}$ refers to the pretrained segmentation model.
With the invisible parts of the vehicle recovered in the input image, the occluded vehicle in the image will appear on the foreground. However, there may exist artifacts in the recovered image $I_r$. Our multi-task framework allows the recovered image to be processed multiple times and finally produces a refined image, e.g., the synthesized image after the second process is produced as: $I^{\prime}_r = G_2(G_1(\mathcal{F}(I_r)))$.
The intuition is based on the correlation between the segmentation completion and the appearance recovery.
In each iteration of the process, the completeness of the recovered segmentation mask affects the quality of the appearance recovery. In the same iteration, the appearance recovery network recovers the appearance while implicitly refining its segmentation mask. Hence, the iterative process may progressively improve the quality of generation, as shown by the example in Fig.~\ref{fig:iters}.

\section{Experiments}

\subsection{Implementation details}

\textbf{3D model pool.} From ShapeNet~\cite{shapenet2015}, we select 401 different classes of vehicles and, for each vehicle, we screenshot each rendered image from 80 different viewpoints. Since the background of the rendered image is very clean, we can simply extract the accurate silhouettes based on thresholds. In this way, we collect 32,080 silhouettes to form the auxiliary 3D model pool.

\textbf{Network structure and training.} In practice, both $G_1$ and $G_2$ downsample the resolution from 256 to 64 and upsample to the original spatial resolution. As the middle layers, there are 8 residual blocks with the dilation rate 2. $D_2$ adopts the Patch-GAN discriminator~\cite{isola2017image}, while $D_{obj}$ and $D_{ins}$ use the same structure except that they have an additional fully connected layer for classification. For hyper-parameters, we set $\lambda=\lambda_1=\lambda_2 = 10$ and $\beta=\beta_1=\beta_2=1$.
In practice, we employ~\cite{he2017mask} as the pretrained segmentation model. 
Our model is implemented in Tensorflow on PC with Intel Core i7-6700 CPU, 32GB RAM, and a single NVIDIA Titan Xp. We first train the segmentation completion network and the appearance recovery network separately using 256$\times$256 images with a batch size of 4, with Adam solver. To train each network, we set the learning rate as ${10^{-4}}$ until the loss plateaus and then lower it to ${10^{-5}}$ until convergence. 
We then train both networks in an end-to-end manner with the learning rate ${10^{-6}}$.

\textbf{Metrics.}
We evaluate our methods in two tasks. For the recovered segmentation mask, we adopt precision, recall, F1-score, Intersection over Union (IoU), the per-pixel $\mathcal{L}_1$ error, and the per-pixel $\mathcal{L}_2$ error as the evaluation metrics. For the recovered appearance of the vehicle, we adopt the per-pixel $\mathcal{L}_1$ error and per-pixel $\mathcal{L}_2$ error. 
Additionally, to evaluate the generation quality, the inception score~\cite{NIPS2016_6125} is often used.
Since we care about the generation quality rather than the diversity, we simply adopt the conditional probability computed by the pretrained Inception~\cite{Szegedy_2016_CVPR} for evaluating the recovered vehicles, denoted as \textit{Inception conditional probability} (ICP).
Besides, similar to the inception score and the FCN-score used in \cite{isola2017image}, we also apply the state-of-the-art segmentation model \cite{he2017mask} trained on Cityscape~\cite{Cordts2016Cityscapes} to segment the whole recovered image with the reference of the ground-truth labels. The intuition is that, if the recovered vehicle in the image is realistic, the segmentation model trained on real images will be able to classify it correctly. Thus, the accuracy of such segmentation for vehicles is adopted as another metric, denoted as \textit{segmentation score} (SS).

\subsection{Occluded Vehicle Dataset} 

To our best knowledge, there is no public dataset providing occluded vehicles with unoccluded segmentation and appearance ground-truth. 
For experiments, we present a new dataset called the Occluded Vehicle Dataset (OVD). We leverage the images of the Cars dataset~\cite{KrauseStarkDengFei-Fei_3DRR2013} as the base images for synthesis. The original dataset contains 16,185 images with 196 classes of cars. Since most cars from the dataset are unoccluded, we manually label the segmentation of the cars as ground truth. Besides, we randomly place some real pedestrians, vehicles, and other objects, which are randomly cropped from the Microsoft COCO dataset \cite{lin2014microsoft} and CityScape \cite{Cordts2016Cityscapes}, as occlusions over the vehicles of these base images. Then, we adopt the Deep Harmonization technique~\cite{Tsai_CVPR_2017} to make these synthetic images look natural. 
In this way, we collect a total of 33,100 images for training and 1000 images for testing. Note that the vehicles in the test images are unseen in the training set. In addition, we also collect and label 100 real-world images as part of the dataset. 
Therefore, we enrich the diversity of the images in four aspects: (1) the number of vehicles; (2) the classes of vehicles; (3) the poses of vehicles; (4) the types of the occlusions. Moreover, we collect and label 4 video sequences (i.e. Vid-1, Vid-2, Vid-3, Vid-4) with occluded vehicles, which are captured from underground parking lots, crowded exhibitions, and streets.
More examples of OVD are shown in the experiment section.

\subsection{Ablation studies}

To analyze our proposed framework, we first evaluate our proposed discriminators in the segmentation completion network, and then evaluate the two-path structure of the appearance recovery network. Lastly, we analyze our iterative generation method. All the ablation studies are performed on the synthetic and real images of the testset in OVD.
\begin{table}
	\centering
	\footnotesize
	\tabcolsep=0.15cm
	\setlength\arrayrulewidth{1.0pt}
	\begin{tabular}{c|c|cccc}
		\toprule
		Structure & Input & L1 $\downarrow$ & L2 $\downarrow$ & F1 $\uparrow$ & IoU $\uparrow$\\
		\midrule
		$D_{standard}$ & \multirow{2}{*}{Syn.} & 0.0702 & 0.0667 & 0.8403 & 0.7421 \\
		$\{D_{obj}, D_{ins}\}$ & & {\bf 0.0559} & {\bf 0.0535} & {\bf 0.8798} & {\bf 0.7939}\\
		\midrule
		$D_{standard}$ & \multirow{2}{*}{Real} & 0.0338 & 0.0335 & 0.8890 & 0.8067 \\
		$\{D_{obj}, D_{ins}\}$ & & {\bf 0.0322} & {\bf 0.0314} & {\bf 0.8981} & {\bf 0.8193}\\
		\midrule
		\midrule
		Structure & Input & L1 $\downarrow$ & L2 $\downarrow$ & ICP $\uparrow$ & SS $\uparrow$\\
		\midrule
		one-path & \multirow{2}{*}{Syn.} & 0.0421 & 0.0181 & 0.6214 & 0.8350 \\
		two-path & & {\bf 0.0364} & {\bf 0.0161} & {\bf 0.6676} & {\bf 0.9411} \\
		\midrule
		one-path & \multirow{2}{*}{Real} & 0.0201 & 0.0077 & 0.8058 & 0.9131 \\
		two-path & & {\bf 0.0171} & {\bf 0.0063} & {\bf 0.8216} & {\bf 0.9292} \\
		\bottomrule
	\end{tabular}
	\caption{\small Ablation experiments for our architectures.}
	\vspace{-0.3cm}
	\label{tab:ab1}
\end{table}

\begin{table}
	\centering
	\footnotesize
	\tabcolsep=0.15cm
	\setlength\arrayrulewidth{1.0pt}
	\begin{tabular}{c|c|cccc}
		\toprule
		Iterations & Input & L1 $\downarrow$ & L2 $\downarrow$ & F1 $\uparrow$ & IoU $\uparrow$\\
		\midrule
		1 & \multirow{3}{*}{Syn.} & 0.0559 & 0.0535 & 0.8798 & 0.7939 \\
		2 & & {\bf 0.0499} & {\bf 0.0480} & {\bf 0.8935} & {\bf 0.8137}\\
		3 & & 0.0510 & 0.0493 & 0.8902 & 0.8080\\
		\midrule
		1 & \multirow{3}{*}{Real} & 0.0322 & {\bf 0.0314} & 0.8890 & {0.8066} \\
		2 & & {\bf 0.0320} & {\bf 0.0314} & {\bf 0.8898} & {\bf 0.8067}\\
		3 & & 0.0342 & 0.0336 & 0.8810 & 0.7926\\
		\midrule
		\midrule
		Iterations & Input & L1 $\downarrow$ & L2 $\downarrow$ & ICP $\uparrow$ & SS $\uparrow$\\
		\midrule
		1 & \multirow{3}{*}{Syn.} & 0.0364 & 0.0161 & 0.6676 & 0.9411 \\
		2 & & {\bf 0.0341} & {\bf 0.0146} & {\bf 0.6765} &  {\bf 0.9545}\\
		3 & & 0.0343 & 0.0148 & 0.6748 & 0.8458\\
		\midrule
		1 & \multirow{3}{*}{Real} & {\bf 0.0171} & {\bf 0.0063} & 0.8216 & 0.9292 \\
		2 & & 0.0173 & {\bf 0.0063} & {\bf 0.8350} & {\bf 0.9356} \\
		3 & & 0.0178 & 0.0066 & 0.8206 & 0.9314 \\
		\bottomrule
	\end{tabular}
	\caption{\small Ablation experiments for studying the iterations of our model in the tasks of segmentation and appearance recovery.}
	\vspace{-0.3cm}
	\label{tab:ab2}
\end{table}

\begin{figure}
	\hspace{-.4cm}
	\setlength{\tabcolsep}{3pt}
	\begin{tabular}{ccc}
		\includegraphics[width=0.16\textwidth,trim={0 1.3cm 0 1.6cm},clip]{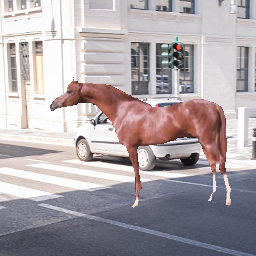}&
		\includegraphics[width=0.16\textwidth,trim={0 1.3cm 0 1.6cm},clip]{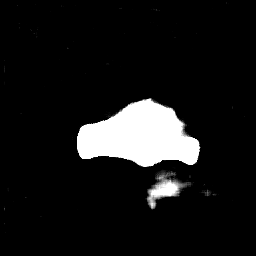}&
		\includegraphics[width=0.16\textwidth,trim={0 1.3cm 0 1.6cm},clip]{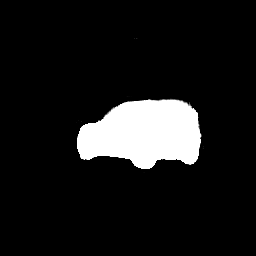}\\
		\includegraphics[width=0.16\textwidth]{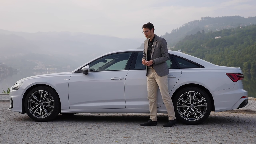}&
		\includegraphics[width=0.16\textwidth]{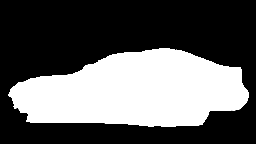}&
		\includegraphics[width=0.16\textwidth]{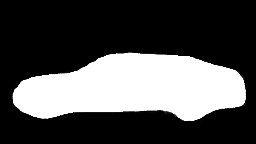}\\
		{\footnotesize Input image}  & {\footnotesize $D_{standard}$} & {\footnotesize $\{D_{obj}, D_{ins}\}$}\\
	\end{tabular}
	\caption{\small Generated complete segmentation masks on exemplar synthetic and real images for evaluating our discriminators. }  
	\vspace{-0.3cm}
	\label{fig:no3d} 
\end{figure}

By evaluating the discriminators in the segmentation completion network, we compare our proposed model ($\{D_{obj}, D_{ins}\}$) against the standard discriminator $D_{standard}$ that is a single discriminator network for real/fake classification. 
As illustrated in Tab.~\ref{tab:ab1}, the quality of segmentation completion from our proposed model is generally improved. As shown in Fig.~\ref{fig:no3d}, the completed segmentation masks may be coarse and noisy using $D_{standard}$.

\begin{figure}
	\hspace{-.4cm}
	\setlength{\tabcolsep}{3pt}
	\begin{tabular}{ccc}
		\includegraphics[width=0.16\textwidth,trim={0 1.3cm 0 1.6cm},clip]{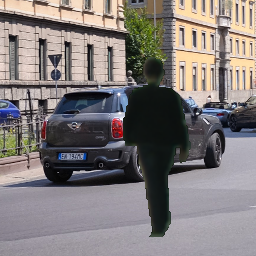}&
		\includegraphics[width=0.16\textwidth,trim={0 1.3cm 0 1.6cm},clip]{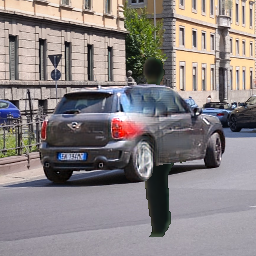}&
		\includegraphics[width=0.16\textwidth,trim={0 1.3cm 0 1.6cm},clip]{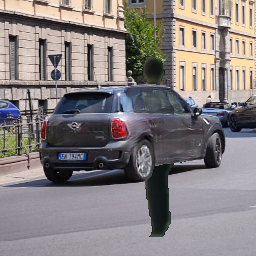}\\
		\includegraphics[width=0.16\textwidth]{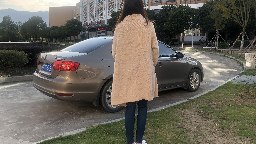}&
		\includegraphics[width=0.16\textwidth]{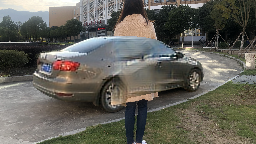}&
		\includegraphics[width=0.16\textwidth]{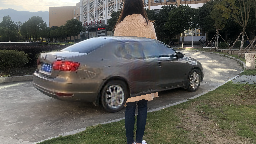}\\
		{\footnotesize Input image}  & {\footnotesize One-path structure} & {\footnotesize Two-path structure}\\
	\end{tabular}
	\caption{\small Recovered appearance on exemplar synthetic and real images for evaluating our proposed two-path structure. }  
	\label{fig:paths} 
	\vspace{-0.3cm}
\end{figure}

To demonstrate the effectiveness of the two-path structure, we compare our two-path structure with the one-path structure which contains the first path only. The second path requires the ground-truth labels, so it cannot be applied in test solely. As illustrated in Tab.~\ref{tab:ab1}, the two-path structure shows the obvious advantages over its counterpart, as the one-path may not be capable enough to fully recover the appearance from the invisible parts, as shown in Fig.\ref{fig:paths}.

\begin{figure}
	\hspace{-.2cm}
	\setlength{\tabcolsep}{3pt}
	\begin{tabular}{ccc}
		\includegraphics[width=0.15\textwidth,trim={0 1.1cm 0 0.3cm},clip]{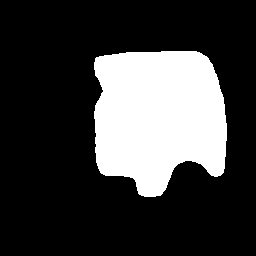}&
		\includegraphics[width=0.15\textwidth,trim={0 1.1cm 0 0.3cm},clip]{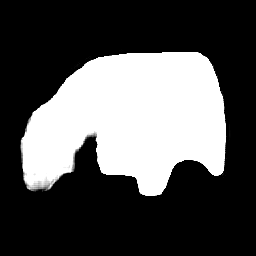}&
		\includegraphics[width=0.15\textwidth,trim={0 1.1cm 0 0.3cm},clip]{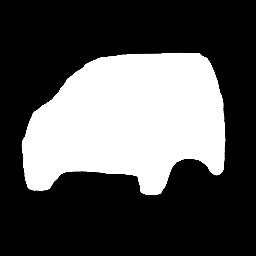}\\
		\includegraphics[width=0.15\textwidth,trim={0 1.1cm 0 0.3cm},clip]{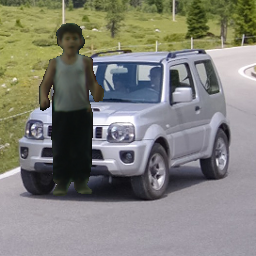}&
		\includegraphics[width=0.15\textwidth,trim={0 1.1cm 0 0.3cm},clip]{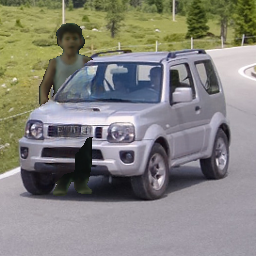}&
		\includegraphics[width=0.15\textwidth,trim={0 1.1cm 0 0.3cm},clip]{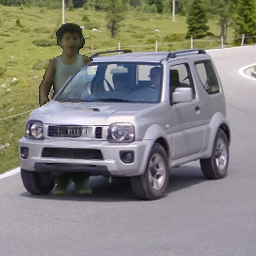}\\
		{\footnotesize Input}  & {\footnotesize $1^{st}$ iteration} & {\footnotesize $2^{nd}$ iteration}\\
	\end{tabular}
	\caption{\small Illustration of an example on our iterative refinement. The first column refers the input image and its corresponding incomplete mask. The second and third column refer to the results produced at the first and the second iteration, respectively.}  
	\label{fig:iters} 
	\vspace{-0.3cm}
\end{figure}

Lastly, we analyze the performance of running our model for 1, 2, and 3 iterations in the tasks of segmentation completion and appearance recovery. In specific, the model running for 1 iteration refers to the process of passing input images through two generators once.
Generally, we obtain the optimal performance in the second iteration. For synthetic images, due to different kinds of synthetic occlusions in images, our model requires multiple iterations to progressively remove the occlusions and recover the missing details. We show an example of the progressive refinement in Fig.~\ref{fig:iters}.
For real images with less severe occlusions, the second iteration only slightly improves the performance of recovering segmentation and appearance, since the model on the first iteration has already produced recognizable shapes. But its improvement on ICP indicates that the iterative process still manages to refine the appearance for the recovered object. 

\begin{table*}
	\centering
	\footnotesize
	\tabcolsep=0.13cm
	\setlength\arrayrulewidth{1.0pt}
	\begin{tabular}{c||cccccc||cccccc}
		\toprule
		& \multicolumn{6}{c||}{\textit{Synthetic images}} & \multicolumn{6}{c}{\textit{Real images}}\\
		Model & Prec. $\uparrow$ & Recall $\uparrow$ & F1 $\uparrow$ & IoU $\uparrow$ & L1 $\downarrow$ & L2 $\downarrow$ & Prec. $\uparrow$ & Recall $\uparrow$ & F1 $\uparrow$ & IoU $\uparrow$ & L1 $\downarrow$ & L2 $\downarrow$\\
		\midrule
		Mask R-CNN~\cite{he2017mask} 
		& 0.7803 & 0.7914 & 0.7653 & 0.6246 & 0.1174 & 0.1162 & 0.7066	& 0.9197 & 0.7953	& 0.6619	& 0.0731	& 0.0730\\
		\rowcolor[gray]{0.8}Deeplab~\cite{chen2018deeplab} &0.8810 &0.8779 &0.8794 &0.7937 &0.0574 &0.0559 & 0.8918 & 0.7508 & 0.8111 & 0.6976 & 0.0545 & 0.0544\\
		SharpMask~\cite{pinheiro2016learning} & 0.8286 & \fir{\textbf{0.9404}} & 0.8751 & 0.7840 &0.0646 & 0.0635 & 0.8025 & \fir{\textbf{0.9518}} & 0.8669 & 0.7693 & 0.0463 & 0.0462\\ 
		\rowcolor[gray]{0.8} pix2pix~\cite{isola2017image}	
		& 0.8865 &	0.8906 &	\se{\textbf{0.8821}}	& 0.7932	& 0.0575 &	0.0557 & 0.8471	& 0.9055 & 0.8718 &	0.7763 &	0.0414 &	\se{\textbf{0.0407}}\\
		SeGAN~\cite{ehsani2018segan}	
		& 0.7931	& \se{\textbf{0.9016}} &	0.8367 &	0.7236 &	0.0846	& 0.0835 & 0.7477	& \se{\textbf{0.9417}} & 0.8303 &	0.7133	& 0.0603 &	0.0602\\
		\rowcolor[gray]{0.8} Ours ($1^{st}$ iter.)	
		& \se{\textbf{0.9590}}	& 0.8229	& 0.8798	& \se{\textbf{0.7939}}	& \se{\textbf{0.0559}}	& \se{\textbf{0.0535}} & \se{\textbf{0.9821}} &	0.8176 & \se{\textbf{0.8890}}	& \fir{\textbf{0.8067}}	& \se{\textbf{0.0322}}	& \fir{\textbf{0.0314}}\\
		Ours ($2^{nd}$ iter.)	
		& \fir{\textbf{0.9625}}	& 0.8416	& \fir{\textbf{0.8935}}	& \fir{\textbf{0.8137}}	& \fir{\textbf{0.0499}}	& \fir{\textbf{0.0480}} & \fir{\textbf{0.9854}}	& 0.8148 & \fir{\textbf{0.8898}}	& \se{\textbf{0.8066}}	& \fir{\textbf{0.0320}} &	\fir{\textbf{0.0314}}\\
		\bottomrule
	\end{tabular}
	\caption{\small The comparison results of segmentation completion in Occluded Vehicle dataset. On each column, the top performer is marked in red while the second one is marked in blue.}
	\vspace{-3mm}
	\label{tab:seg}
\end{table*}

\subsection{Results analysis}

We compare our model with the state-of-the-art methods in two tasks, i.e. segmentation completion and appearance recovery. For the task of segmentation completion, we compare with Mask R-CNN~\cite{he2017mask}, Deeplab~\cite{chen2018deeplab}, SharpMask~\cite{pinheiro2016learning}, pix2pix~\cite{isola2017image}, and SeGAN~\cite{ehsani2018segan}. For appearance recovery, we compare with Deepfill~\cite{yu2018generative}, pix2pix, and SeGAN. Specifically, Mask R-CNN and Deeplab are the state-of-the-art segmentation models. SharpMask has been proposed to complete and refine the generated masks. 
As a supervised GAN model, pix2pix can be applied in several applications, including segmentation and image synthesis. 
Deepfill is one of the state-of-the-art GAN-based inpainting methods and SeGAN claims to achieve the state-of-the-art performance in both amodal segmentation and appearace recovery. 
In experiments, all of them are fine-tuned with our training data for our tasks. Besides, we run our model for 1 iteration and 2 iterations respectively for comparison. The evaluations are separately performed on the synthetic images and the real images of our dataset.

\begin{figure*}

\setlength{\tabcolsep}{2pt}
\hspace{-.5cm}
\begin{tabular}{ccccccc}
	
\includegraphics[width=0.145\textwidth,trim={0 2cm 0 1cm},clip]{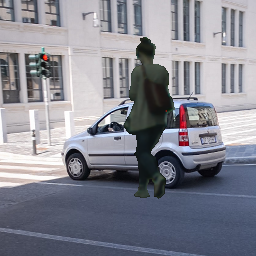}&
\includegraphics[width=0.145\textwidth,trim={0 2cm 0 1cm},clip]{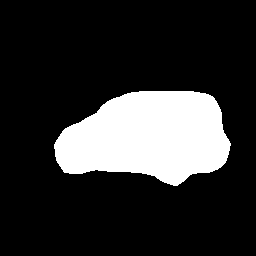}&
\includegraphics[width=0.145\textwidth,trim={0 2cm 0 1cm},clip]{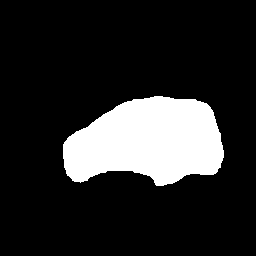}&
\includegraphics[width=0.145\textwidth,trim={0 2cm 0 1cm},clip]{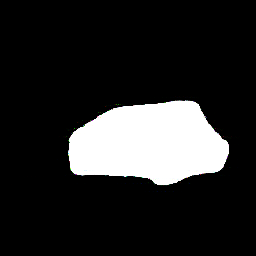}&
\includegraphics[width=0.145\textwidth,trim={0 2cm 0 1cm},clip]{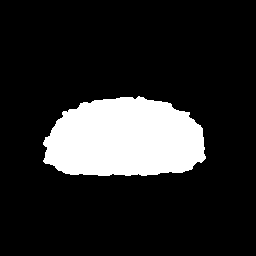}&
\includegraphics[width=0.145\textwidth,trim={0 2cm 0 1cm},clip]{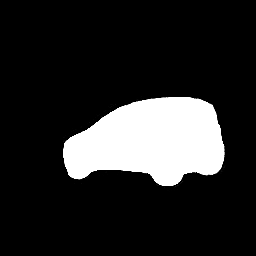}&
\includegraphics[width=0.145\textwidth,trim={0 2cm 0 1cm},clip]{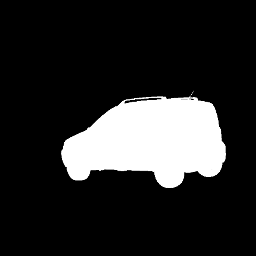}\\	
\includegraphics[width=0.145\textwidth,trim={0 2cm 0 2cm},clip]{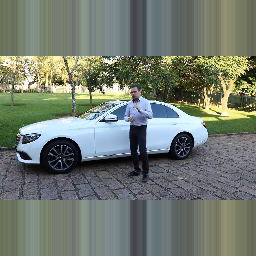}&
\includegraphics[width=0.145\textwidth,trim={0 2cm 0 2cm},clip]{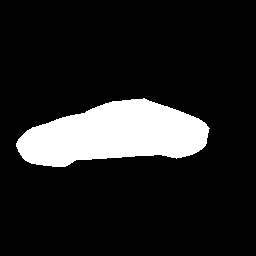}&
\includegraphics[width=0.145\textwidth,trim={0 2cm 0 2cm},clip]{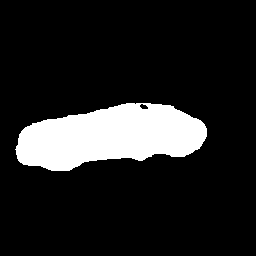}&
\includegraphics[width=0.145\textwidth,trim={0 2cm 0 2cm},clip]{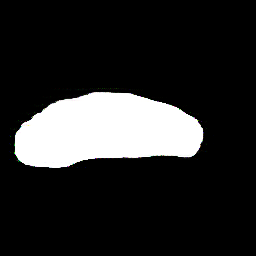}&
\includegraphics[width=0.145\textwidth,trim={0 2cm 0 2cm},clip]{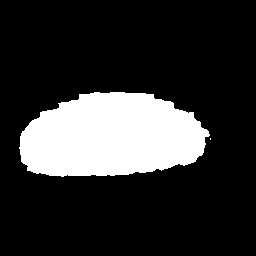}&
\includegraphics[width=0.145\textwidth,trim={0 2cm 0 2cm},clip]{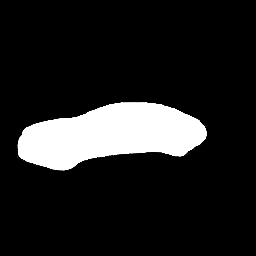}&
\includegraphics[width=0.145\textwidth,trim={0 2cm 0 2cm},clip]{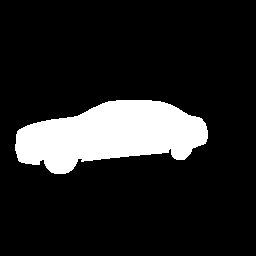}\\
\includegraphics[width=0.145\textwidth,trim={0 2cm 0 2cm},clip]{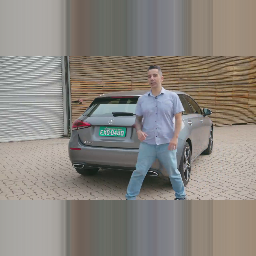}&
\includegraphics[width=0.145\textwidth,trim={0 2cm 0 2cm},clip]{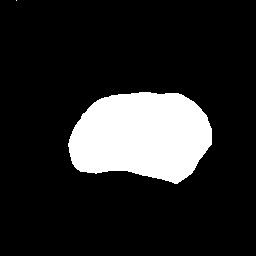}&
\includegraphics[width=0.145\textwidth,trim={0 2cm 0 2cm},clip]{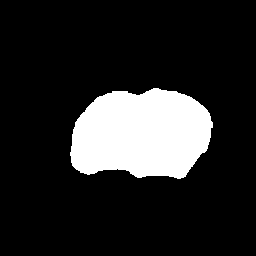}&
\includegraphics[width=0.145\textwidth,trim={0 2cm 0 2cm},clip]{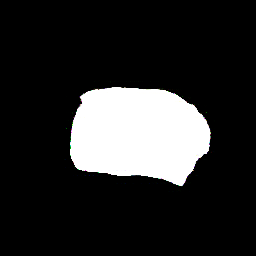}&
\includegraphics[width=0.145\textwidth,trim={0 2cm 0 2cm},clip]{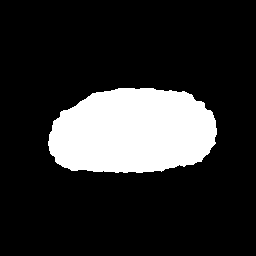}&
\includegraphics[width=0.145\textwidth,trim={0 2cm 0 2cm},clip]{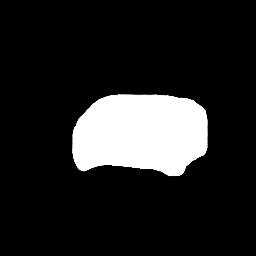}&
\includegraphics[width=0.145\textwidth,trim={0 2cm 0 2cm},clip]{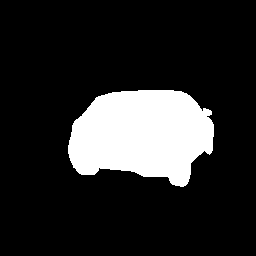}\\
\includegraphics[width=0.145\textwidth,trim={0 2cm 0 2cm},clip]{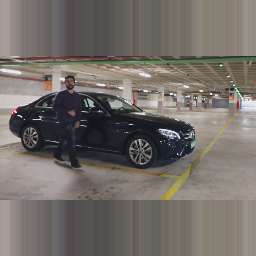}&
\includegraphics[width=0.145\textwidth,trim={0 2cm 0 2cm},clip]{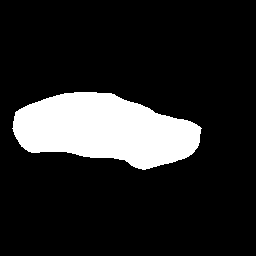}&
\includegraphics[width=0.145\textwidth,trim={0 2cm 0 2cm},clip]{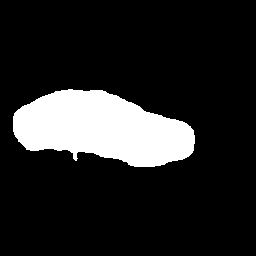}&
\includegraphics[width=0.145\textwidth,trim={0 2cm 0 2cm},clip]{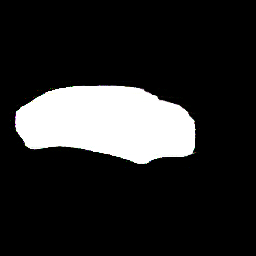}&
\includegraphics[width=0.145\textwidth,trim={0 2cm 0 2cm},clip]{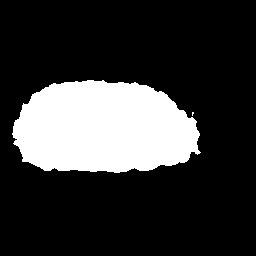}&
\includegraphics[width=0.145\textwidth,trim={0 2cm 0 2cm},clip]{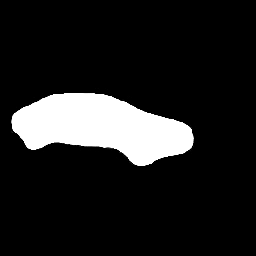}&
\includegraphics[width=0.145\textwidth,trim={0 2cm 0 2cm},clip]{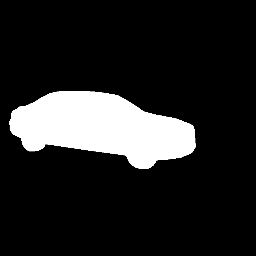}\\
{\footnotesize Input}  & {\footnotesize Deeplab~\cite{chen2018deeplab}} & {\footnotesize SharpMask~\cite{pinheiro2016learning}} & {\footnotesize pix2pix~\cite{isola2017image}} & {\footnotesize SeGAN~\cite{ehsani2018segan}} & {\footnotesize Ours} & {\footnotesize Ground truth}\\
\end{tabular}
    \caption{\small Examples of the segmentation completion comparison. }
    \vspace{-0.3cm}
    \label{fig:seg_comp} 
\end{figure*}

For segmentation completion, we demonstrate the comparison results in Tab. \ref{tab:seg}. Generally, the result shows that our model with or without the iterative refinement outperforms the prior methods. 
As illustrated in Fig.~\ref{fig:seg_comp}, our model can produce masks with smooth contours and clear shapes of wheels and bodywork, due to the involvement of the object discriminator and the auxiliary 3D model pool. The results of Deeplab are comparable to ours, but the shapes of wheels and bodyworks are not clear. Since SeGAN generates masks with the low-resolution (i.e. $58 \times 58$) and upsamples them to $256 \times 256$, their results appear to be coarse. SharpMask and pix2pix can produce more complete and finer masks than SeGAN, but their contours are not adequately smooth.

\begin{figure*}
	
	\setlength{\tabcolsep}{2pt}
	\hspace{-.5cm}
	\begin{tabular}{ccccccc}
		\includegraphics[width=0.145\textwidth,trim={0 2cm 0 1cm},clip]{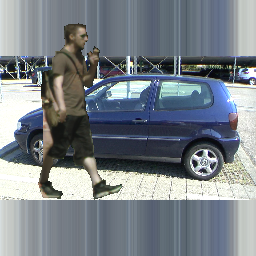}&
		\includegraphics[width=0.145\textwidth,trim={0 2cm 0 1cm},clip]{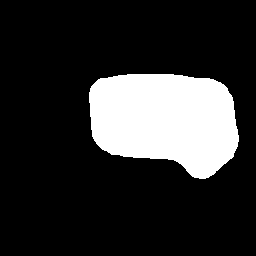}&
		\includegraphics[width=0.145\textwidth,trim={0 2cm 0 1cm},clip]{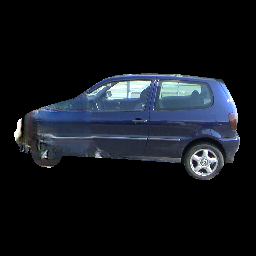}&
		\includegraphics[width=0.145\textwidth,trim={0 2cm 0 1cm},clip]{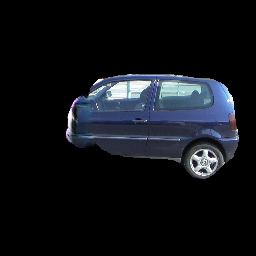}&
		\includegraphics[width=0.145\textwidth,trim={0 2cm 0 1cm},clip]{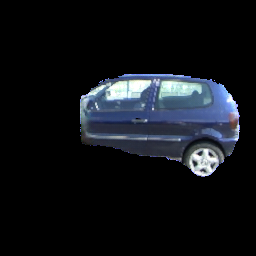}&
		\includegraphics[width=0.145\textwidth,trim={0 2cm 0 1cm},clip]{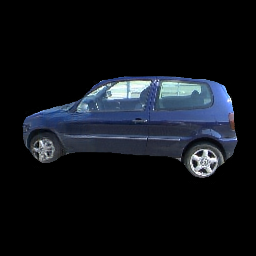}&
		\includegraphics[width=0.145\textwidth,trim={0 2cm 0 1cm},clip]{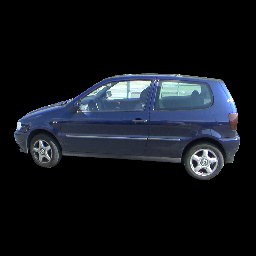}\\
		\includegraphics[width=0.145\textwidth,trim={0 2cm 0 2cm},clip]{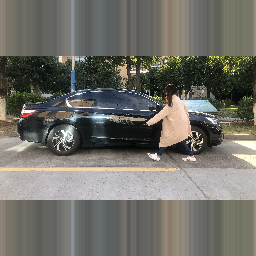}&
		\includegraphics[width=0.145\textwidth,trim={0 2cm 0 2cm},clip]{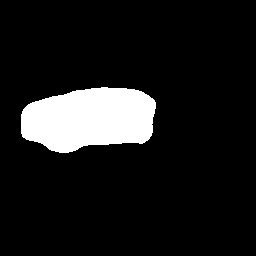}&
		\includegraphics[width=0.145\textwidth,trim={0 2cm 0 2cm},clip]{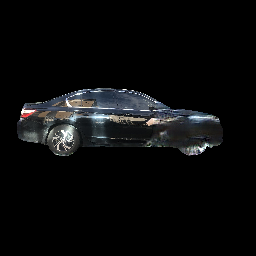}&
		\includegraphics[width=0.145\textwidth,trim={0 2cm 0 2cm},clip]{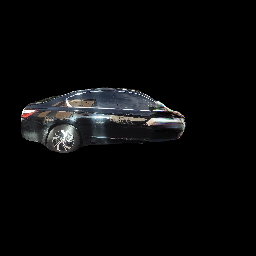}&
		\includegraphics[width=0.145\textwidth,trim={0 2cm 0 2cm},clip]{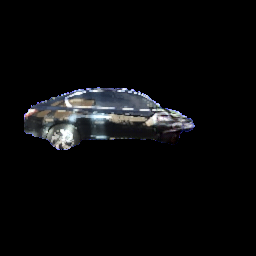}&
		\includegraphics[width=0.145\textwidth,trim={0 2cm 0 2cm},clip]{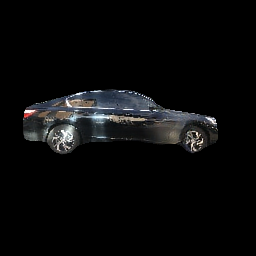}&
		\includegraphics[width=0.145\textwidth,trim={0 2cm 0 2cm},clip]{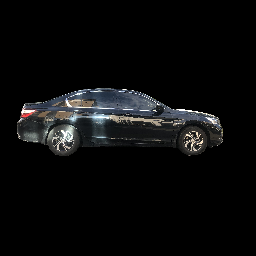}\\
		\includegraphics[width=0.145\textwidth,trim={0 2cm 0 2cm},clip]{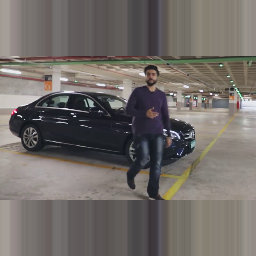}&
		\includegraphics[width=0.145\textwidth,trim={0 2cm 0 2cm},clip]{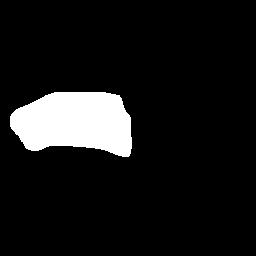}&
		\includegraphics[width=0.145\textwidth,trim={0 2cm 0 2cm},clip]{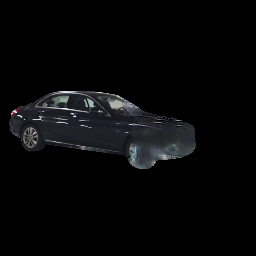}&
		\includegraphics[width=0.145\textwidth,trim={0 2cm 0 2cm},clip]{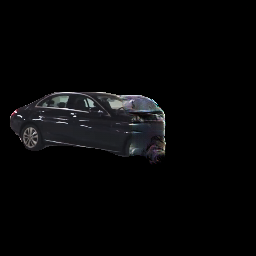}&
		\includegraphics[width=0.145\textwidth,trim={0 2cm 0 2cm},clip]{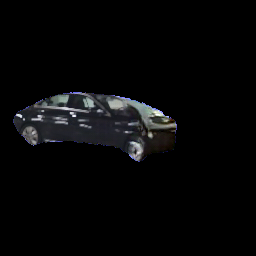}&
		\includegraphics[width=0.145\textwidth,trim={0 2cm 0 2cm},clip]{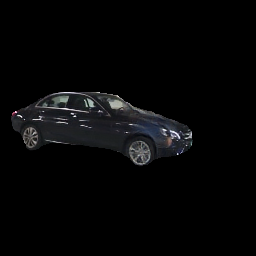}&
		\includegraphics[width=0.145\textwidth,trim={0 2cm 0 2cm},clip]{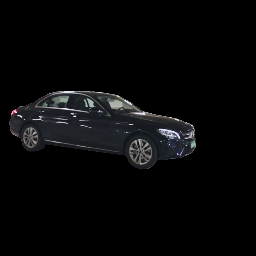}\\
		\includegraphics[width=0.145\textwidth,trim={0 2cm 0 2cm},clip]{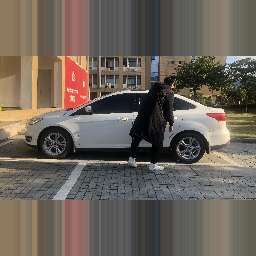}&
		\includegraphics[width=0.145\textwidth,trim={0 2cm 0 2cm},clip]{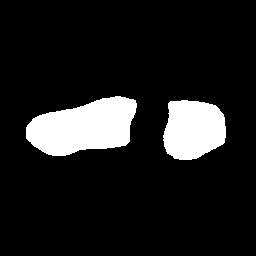}&
		\includegraphics[width=0.145\textwidth,trim={0 2cm 0 2cm},clip]{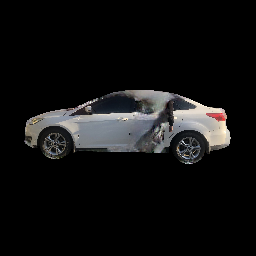}&
		\includegraphics[width=0.145\textwidth,trim={0 2cm 0 2cm},clip]{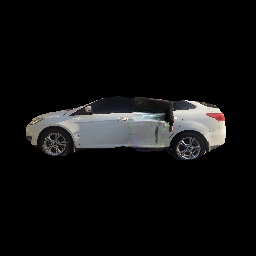}&
		\includegraphics[width=0.145\textwidth,trim={0 2cm 0 2cm},clip]{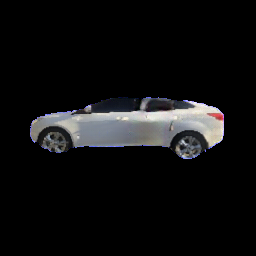}&
		\includegraphics[width=0.145\textwidth,trim={0 2cm 0 2cm},clip]{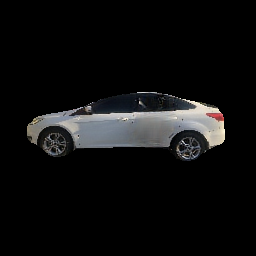}&
		\includegraphics[width=0.145\textwidth,trim={0 2cm 0 2cm},clip]{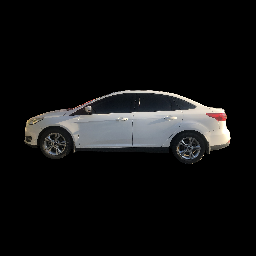}\\	
		{\footnotesize Input}  & {\footnotesize Incomplete mask} & {\footnotesize Deepfill~\cite{yu2018generative}} & {\footnotesize pix2pix~\cite{isola2017image}} & {\footnotesize SeGAN~\cite{ehsani2018segan}} & {\footnotesize Ours} & {\footnotesize Ground truth}\\
	\end{tabular}
	\caption{\small Examples of the appearance recovery comparison.}
	\vspace{-0.3cm}
	\label{fig:app_comp} 
\end{figure*}

\begin{figure*}
	\begin{tabular}{ccccc}
		
		\includegraphics[trim={0cm 1cm 0 2cm},clip,width=0.18\textwidth,height=0.8in]{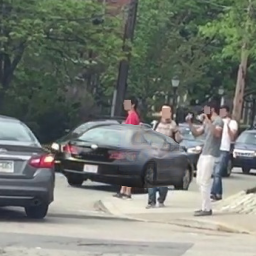}&
		\includegraphics[trim={0cm 0cm 0 2cm},clip,width=0.18\textwidth,height=0.8in]{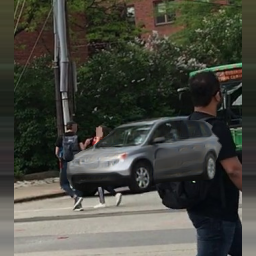}&
		\includegraphics[trim={0cm 0cm 0 3cm},clip,width=0.18\textwidth,height=0.8in]{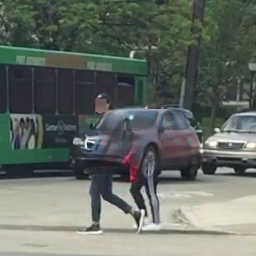}&
		\includegraphics[trim={0cm 1cm 0 1cm},clip,width=0.18\textwidth,height=0.8in]{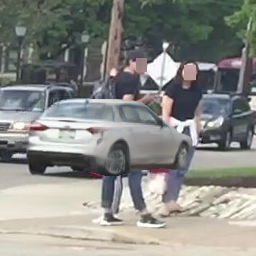}&
		\includegraphics[trim={0cm 2cm 0 2cm},clip,width=0.18\textwidth,height=0.8in]{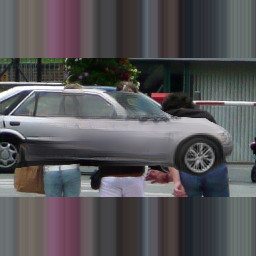}\\
		\includegraphics[trim={0cm 1cm 0 1cm},clip,width=0.18\textwidth,height=0.8in]{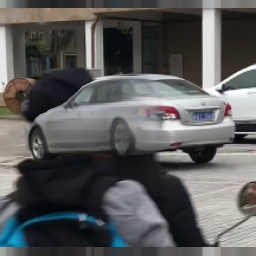}&
		\includegraphics[trim={0cm 1cm 0 3cm},clip,width=0.18\textwidth,height=0.8in]{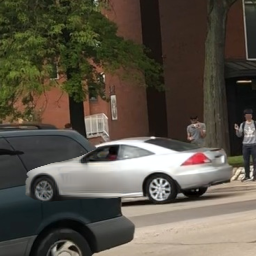}&
		\includegraphics[trim={1cm 2cm 0 2cm},clip,width=0.18\textwidth,height=0.8in]{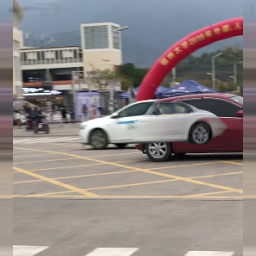}&
		\includegraphics[trim={0cm 1cm 0 2cm},clip,width=0.18\textwidth,height=0.8in]{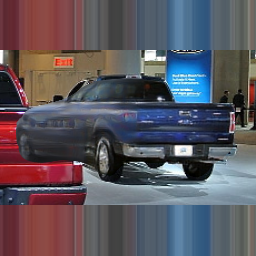}&
		\includegraphics[trim={0cm 1cm 0 1cm},clip,width=0.18\textwidth,height=0.8in]{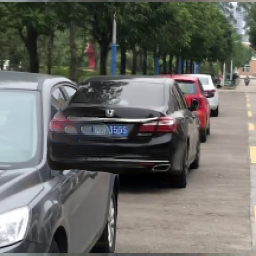}\\
		\includegraphics[trim={0cm 1cm 0 1cm},clip,width=0.18\textwidth,height=0.8in]{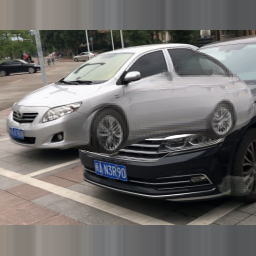}&
		\includegraphics[trim={0cm 1cm 0 2cm},clip,width=0.18\textwidth,height=0.8in]{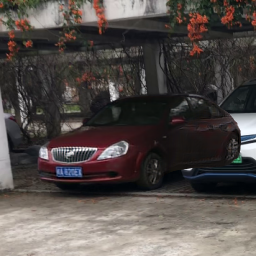}&
		\includegraphics[trim={0cm 2cm 0 2cm},clip,width=0.18\textwidth,height=0.8in]{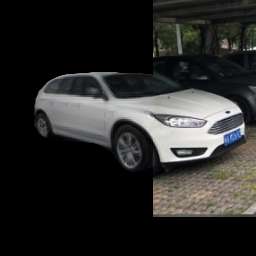}&
		\includegraphics[trim={0cm 1cm 0 1cm},clip,width=0.18\textwidth,height=0.8in]{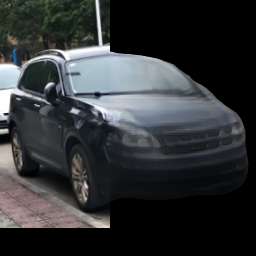}&
		\includegraphics[trim={1cm 2cm 0 2cm},clip,width=0.18\textwidth,height=0.8in]{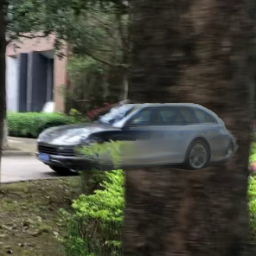}\\
	\end{tabular}
	\caption{\footnotesize Recovered real occluded vehicles from other public datasets, including vehicle-occluding-vehicle, multi-person-occluding-vehicle, and truncated vehicles.}
	\label{fig:extra}
\end{figure*}

For appearance recovery, the comparison results are shown in Tab. \ref{tab:app}. Since Deepfill requires the image context without vehicles, we provide the ground-truth segmentation mask $M^{gt}$. For the fair comparison, we also provide the ground-truth masks for the other methods as well. As observed, our method demonstrates superior performance over others. However, since the ground-truth masks are provided, the iterative refinement does not show much effect and even degrades the model performance a little. 
In addition, we perform comparisons of appearance recovery for methods given their predicted masks. These comparisons are performed in both synthetic and real images. 
According to ICP for evaluating the recovered vehicles and SS for evaluating the recovered image, our method generates more plausible images. As shown in Fig.~\ref{fig:app_comp}, only Deepfill is provided with the ground-truth mask, so its recovered vehicles have more complete shape but without much details. The other comparison results are generated based on the input images or their predicted masks. Due to our appearance recovery network, we can paint more details on the invisible parts. Taking the first and the third row in Fig.~\ref{fig:app_comp} as examples, our model manages to paint a wheel on the proper position in the image. More results are illustrated in Fig.~\ref{fig:extra} that recovers occluded vehicles from other public datasets, e.g. COCO. 

\begin{table}
	\centering
	\footnotesize
	\tabcolsep=0.1cm
	\setlength\arrayrulewidth{1.0pt}
	\begin{tabular}{c|c|c|caca}
		\toprule
		Model & Type & Input & L1 $\downarrow$ & L2 $\downarrow$ & ICP $\uparrow$ & SS $\uparrow$\\
		\midrule
		Deepfill~\cite{yu2018generative} &  & & 0.0284 &	0.0107	& 0.5620 & 0.8295\\
		pix2pix~\cite{isola2017image} & &  & 0.0174	& 0.0060	& 0.7081 & 0.9410\\
		SeGAN~\cite{ehsani2018segan} & Syn. & ${M}^{gt}$ &	0.0181&	0.0055	& 0.6662 & 0.9371\\
		Ours ($1^{st}$ iter.)	& & & \se{\textbf{0.0159}}	& \fir{\textbf{0.0038}} &	\fir{\textbf{0.7436}} & \fir{\textbf{0.9458}}\\
		Ours ($2^{nd}$ iter.)	& & & \fir{\textbf{0.0158}}	& \se{\textbf{0.0039}}	& \se{\textbf{0.7267}} & \se{\textbf{0.9447}}\\
		
		\midrule
		pix2pix~\cite{isola2017image} & \multirow{4}{*}{Syn.} & \multirow{4}{*}{$M$}& 0.0455	& 0.0226	& 0.6337 & 0.8825\\
		SeGAN~\cite{ehsani2018segan} & & &	0.0499 &	0.0224	& 0.6138 & 0.9165\\
		Ours ($1^{st}$ iter.)	& & & \se{\textbf{0.0364}} &	\se{\textbf{0.0161}} &	\se{\textbf{0.6676}} & \se{\textbf{0.9411}}\\
		Ours ($2^{nd}$ iter.)	& & & \fir{\textbf{0.0341}} &	\fir{\textbf{0.0146}} &	\fir{\textbf{0.6765}} & \fir{\textbf{0.9545}}\\
		\midrule
		pix2pix~\cite{isola2017image} & \multirow{4}{*}{Real} &\multirow{4}{*}{$M$} & 0.0182	& \se{\textbf{0.0074}}	& 0.7888 & 0.9165\\
		SeGAN~\cite{ehsani2018segan} & & &	0.0256 &	0.0114	& 0.4984 & 0.9192\\
		Ours ($1^{st}$ iter.)	& & & \fir{\textbf{0.0171}} &	\fir{\textbf{0.0063}}	& \se{\textbf{0.8216}} & \se{\textbf{0.9292}}\\
		Ours ($2^{nd}$ iter.)	& & & \se{\textbf{0.0173}} &	\fir{\textbf{0.0063}} &	\fir{\textbf{0.8350}} & \fir{\textbf{0.9356}}\\
		\bottomrule
	\end{tabular}
	\caption{\small The comparison results of Appearance recovery for the synthetic and real images. In order to perform fair comparisons, we assign the ground-truth segmentation mask $M^{gt}$ and the predicted recovered segmentation mask $M$ as the inputs, respectively. }
	\label{tab:app}
\end{table}


\subsection{User study}

To demonstrate the generation quality of our recovered occluded vehicle, we perform a human user study with 12 participants including 7 males and 5 females. 100 randomly selected images from our dataset and their corresponding images recovered from different methods are anonymously shown for each participant in random order. We let the participants to select the best recovered vehicle among the generated results of the approaches including Deepfill~\cite{yu2018generative}, pix2pix~\cite{isola2017image}, SeGAN~\cite{ehsani2018segan}, and ours. Following the comparison settings of our experiments, except that Deepfill is provided with the ground-truth mask as input, the other methods that have been fine-tuned produce results based on the input image or their own predicted masks. 
In Tab.~\ref{tab:user}, we show how often the result of a particular method is chosen as the best generated image. As observed, 
our approach outperforms others in the user study. Since Deepfill generates results based on ground-truth mask, it achieves better results than pix2pix and SeGAN. 

\begin{table}
	\centering
	\small
	\tabcolsep=0.15cm
	\setlength\arrayrulewidth{1.0pt}
	\begin{tabular}{c||cccc}
		\toprule
		Method & Deepfill~\cite{yu2018generative} & pix2pix~\cite{isola2017image} & SeGAN~\cite{ehsani2018segan} & Ours\\
		\midrule
		Chosen as best & 10.9\% & 1.8\% & 2.5\% & 84.8\% \\
		\bottomrule
	\end{tabular}
	\caption{\small User study results. We illustrate how often the participants in user study choose the results of our approach as the best against the state-of-the-art approaches.}
	\label{tab:user}
\end{table}

\subsection{Application for occluded vehicle tracking}

We apply our method in four real-world videos (Vid-1, Vid-2, Vid-3, Vid-4) to recover the vehicles to be unoccluded. Then, we apply the same tracker KCF~\cite{henriques2015tracking} to track the vehicles from the original videos and the recovered videos, respectively. The results are illustrated in Tab.~\ref{tab:track} in terms of average pixel error (APE) and average overlap (AO). The results indicate that our recovered videos benefit the vehicle tracking.
The examples in Fig.~\ref{fig:teaser} from Vid-1 and Vid-3 show the segmentation mask and the appearance are well recovered to assist tracking under occlusions.

\begin{table}
	\centering                          
	\footnotesize
	\tabcolsep=0.15cm
	\setlength\arrayrulewidth{1.0pt}
	\begin{tabular}{c||cc|cc}
		\toprule
		& \multicolumn{2}{c|}{APE $\downarrow$} & \multicolumn{2}{c}{AO $\uparrow$} \\
		& Original & Recovered & Original & Recovered\\
		\midrule
		Vid-1 & 34.60 & {\bf 8.32} & 0.6489 & {\bf 0.8072}\\
		Vid-2 & 26.30 & {\bf 15.83} & 0.7285 & {\bf 0.8040}\\
		Vid-3 & 87.71 & {\bf 21.51} & 0.3584 & {\bf 0.6755}\\
		Vid-4 & 7.29  & {\bf 5.67} & 0.7494 & {\bf 0.7497}\\
		\bottomrule
	\end{tabular}
	\caption{\small Tracking performance comparison for the original and the recovered videos. The better number in comparison is in bold.}
	\label{tab:track}
\end{table}

\subsection{Conclusion} 

In this paper, we propose an iterative multi-task framework to recover the segmentation mask and the appearance for occluded vehicles. In particular, we propose two coupled discriminators and a two-path structure with a shared network to accomplish this purpose. We evaluate our method in a proposed dataset and demonstrate its state-of-the-art performance. Moreover, we show our method can benefit the occluded vehicle tracking.

\ifCLASSOPTIONcaptionsoff
  \newpage
\fi



%

%


{\small
	\bibliographystyle{ieee}
	\bibliography{egbib}
}




\end{document}